# Deep Learning in Robotics: A Review of Recent Research

Harry A. Pierson (corresponding author)

*Department of Industrial Engineering, University of Arkansas, Fayetteville, AR, USA*

4207 Bell Engineering Center

1 University of Arkansas

Fayetteville, AR  72701

hapierso@uark.edu

+1 (479) 575-6034

Michael S. Gashler

*Department of Computer Science and Computer Engineering, University of Arkansas, Fayetteville, AR, USA*

504 J. B. Hunt Building

1 University of Arkansas

Fayetteville, AR  72701

mgashler@uark.edu

# Deep Learning in Robotics: A Review of Recent Research

Advances in deep learning over the last decade have led to a flurry of research in the application of deep artificial neural networks to robotic systems, with at least thirty papers published on the subject between 2014 and the present. This review discusses the applications, benefits, and limitations of deep learning vis-à-vis physical robotic systems, using contemporary research as exemplars. It is intended to communicate recent advances to the wider robotics community and inspire additional interest in and application of deep learning in robotics.

Keywords: deep neural networks; artificial intelligence; human-robot interaction

## 1. Introduction

Deep learning is the science of training large artificial neural networks. Deep neural networks (DNNs) can have hundreds of millions of parameters [1, 2], allowing them to model complex functions such as nonlinear dynamics. They form compact representations of state from raw, high-dimensional, multimodal sensor data commonly found in robotic systems [3], and unlike many machine learning methods, they do not require a human expert to hand-engineer feature vectors from sensor data at design time. DNNs can, however, present particular challenges in physical robotic systems, where generating training data is generally expensive, and sub-optimal performance in training poses a danger in some applications. Yet, despite such challenges, roboticists are finding creative alternatives, such as leveraging training data via digital manipulation, automating training, and employing multiple DNNs to improve performance and reduce training time.

Applying deep learning to robotics is an active research area, with at least thirty papers published on the subject from 2014 through the time of this writing. This review presents a summary of this recent research with particular emphasis on the benefits and challenges vis-à-vis robotics. A primer on deep learning is followed by a discussion of how common DNN structures are used in robotics and in examples from the recent literature. Practical considerations for roboticists wishing to use DNNs are also provided. Finally, limitations of and strategies that mitigate these as well as future trends are discussed.

## 2. Deep learning

### 2.1 A brief history of deep learning

The basic principles of linear regression were used by Gauss and Legendre [4], and many of those same principles still cover what researchers in deep learning study. However, several important advances have slowly transformed regression into what we now call deep learning. First, the addition of an activation function enabled regression methods to fit to nonlinear functions. It also introduced some biological similarity with brain cells [5].

Next, nonlinear models were stacked in "layers" to create powerful models, called *multi-layer perceptrons*. In the 1960s a few researchers independently figured out how to differentiate multi-layer perceptrons [6], and by the 1980s, it evolved into a popular method for training them, called backpropagation [7, 8]. It was soon proven that multi-layer perceptrons were universal function approximators [9], meaning they could fit to any data, no matter how complex, with arbitrary precision, using a finite number of regression units. In many ways, backpropagation marked the beginning of the deep learning revolution; however, researchers still mostly limited their neural networks to a few layers because of *the problem of vanishing gradients* [10, 11]. Deeper neural networks took exponentially longer to train.

Neural networks were successfully applied for robotics control as early as the 1980s [12]. It was quickly recognized that nonlinear regression provided the functionality that was needed for operating dynamical systems in continuous spaces [13, 14], and closely related fuzzy systems seemed well suited for nominal logical control decisions [15]. Even as early as 1989, Pomerleau's ALVINN [16] famously demonstrated that neural networks were effective for helping vehicles to stay in their lanes. However, neural networks were still generally too slow to digest entire images, or perform the complex tasks necessary for many robotics applications.

In the 2000s, researchers began using graphical processing units (GPUs) to parallelize implementations of artificial neural networks [17]. The largest bottleneck in training neural networks is a matrix-vector multiplication step, which can be parallelized using GPUs. In 2006, Hinton presented a training method that he demonstrated to be effective with a many-layered neural network [18]. The near-simultaneous emergence of these technologies triggered the flurry of research interest that is now propelling deep learning forward at an unprecedented rate [19].

As hardware improved, and as neural networks began to become more practical, they were increasingly found to be effective with real robotics applications. In 2004 RNNPB showed that neural networks could self-organize high-level control schema that generalized effectively with several robotics test problems [20]. In 2008, neuroscientists made advances in recognizing how animals achieved locomotion, and were able to extend this knowledge all the way to neural networks for experimental control of robots [21]. In 2011, TNLDR demonstrated that deep neural nets could effectively model both state and dynamics from strictly unsupervised training with raw images of a simulated robot [22]. Another relevant work is Pomerleau's 2012 book surveying applications for neural networks in perception for robot guidance [23].

In hindsight, we see that chess was considered in the early years of artificial intelligence to be representative of human intelligence over machines [24]. After machines beat world-class chess players [25], a new emblematic task was needed to represent the superior capabilities of human intelligence. Visual recognition was largely accepted to be something easy for humans but difficult for machines [26]. But now, with the emergence of deep learning, humans will not be able to claim that as an advantage for much longer. Deep learning has surged ahead of well-established image recognition techniques [27] and has begun to dominate the benchmarks in handwriting recognition [28], video recognition [29], small-image identification [30], detection in biomedical imaging [31-33], and many others. It has even achieved super-human accuracy in several image recognition contests [27, 34, 35]. Perhaps agility or dexterity will be a forthcoming achievement where machines will begin to demonstrate human-like proficiency. If so, it appears that deep neural networks may be the learning model that enables it.

### *2.2 Common DNN structures*

The idea of using machine learning in controlling robots requires humans to be willing to relinquish a degree of control. This can seem counterintuitive at first, but the benefit for doing so is that the system can then begin to learn on its own. This makes the system capable of adapting, and therefore has potential to ultimately make better use of the direction that comes from humans.

DNNs are well suited for use with robots because they are flexible, and can be used in structures that other machine learning models cannot support. Figure 1 diagrams four common structures for using DNNs with robots.

Structure A (in Figure 1) shows a DNN for regressing arbitrary functions. It is typically trained by presenting a large collection of example training pairs: $\{ <\mathbf{x}_1, \mathbf{y}_1>, <\mathbf{x}_2, \mathbf{y}_2>, \dots, <\mathbf{x}_n, \mathbf{y}_n> \}$. An optimization method is applied to minimize the prediction loss. For regression problems, loss is typically measured with sum-squared error, $\sum_{i}^{n}(\mathbf{y}_i - \hat{\mathbf{y}}_i)^2$, and for classification problems it is often measured with cross-entropy, $-\sum_{i}^{n}\mathbf{y}_i \log \hat{y}_i$, particularly when a softmax layer is used for the output layer

of the neural network [36]. Traditionally, the most popular optimization method for

neural networks is stochastic gradient descent [37], but improved methods such as

RMSProp [38] and Adam [39] have recently garnered widespread usage. Some other considerations for training them effectively are given in Section 2.4. After training is finished, novel vectors may be fed in as $\mathbf{x}$ to compute corresponding predictions for $\mathbf{y}$.

Structure B is called an autoencoder [40]. It is one common model for

facilitating "unsupervised learning." It requires two DNNs, called an "encoder" and a

"decoder." In this configuration, only $\mathbf{x}$ needs to be supplied by the user. $\mathbf{s}$ is a

"latent" or internal encoding that the DNN generates. For example, $\mathbf{x}$ might represent images observed by a robot's camera, containing thousands or even millions of values.

The encoder might use convolutional layers, which are known to be effective for digesting images [35, 41, 42]. $\mathbf{s}$ might be a small vector, perhaps only tens of values.

By learning to reduce $\mathbf{x}$ to $\mathbf{s}$, the autoencoder essentially creates its own internal

encoding of "state." It will not necessarily use an encoding that has meaning for

humans, but it will be sufficient for the DNN to approximately reconstruct $\mathbf{x}$. How are

autoencoders useful in robotics? Sometimes, the robot designer may not know exactly

what values are needed by the robot. Autoencoders enable the system to figure that out

autonomously. This becomes especially useful when a hybrid of supervised and

unsupervised learning is used. For example, the user can impose certain values in $\mathbf{s}$

(perhaps, positional coordinates or joint angles) and the DNNs will learn to work with those values, using the other free elements in **s** for its own encoding purposes. Autoencoders may also be used to initialize some parts of Structure C [22]. Generative models are closely related to autoencoders. They utilize just the decoder portion of the model to predict observations from an internal representation of state.

Structure C is a type of "recurrent neural network," which is designed to model dynamic systems, including robots. It is often trained with an approach called "backpropagation through time" [43, 44]. Many advances, such as "long short-term memory units," have made recurrent neural networks much stronger [27, 45]. In this configuration, **u** represents a control signal. **u** may also contain recent observations. **s** is an internal representation of future state, and **x** is a vector of anticipated future observations. The transition function approximates how the control signal will affect state over time. Just as with autoencoders, the representation of state can be entirely latent, or partially imposed by the user. (If it were entirely imposed, the model would be prevented from learning.) If **x** includes an estimate of the utility of state **s**, then this configuration is used in "model-based reinforcement learning" [46].

Structure D learns a control policy. It can facilitate "model-free" reinforcement learning. It uses a DNN to evaluate the utility or quality, **q**, of potential control vectors. **s** is a representation of state, and **u** is a control vector. (Gradient methods can find the values for **u** that maximize **q**. In cases with discrete control vectors, **u** may be omitted from the input-end and **q** augmented to contain an evaluation of each control vector.) Configurations like this are used when an objective task is known for the robot to perform, but the user does not know exactly how to achieve it. By rewarding the robot for accomplishing the task, it can be trained to learn how to prioritize its own choices for actions. As one prominent example, reinforcement learning was used to teach a machine to play a wide range of Atari video games [47].

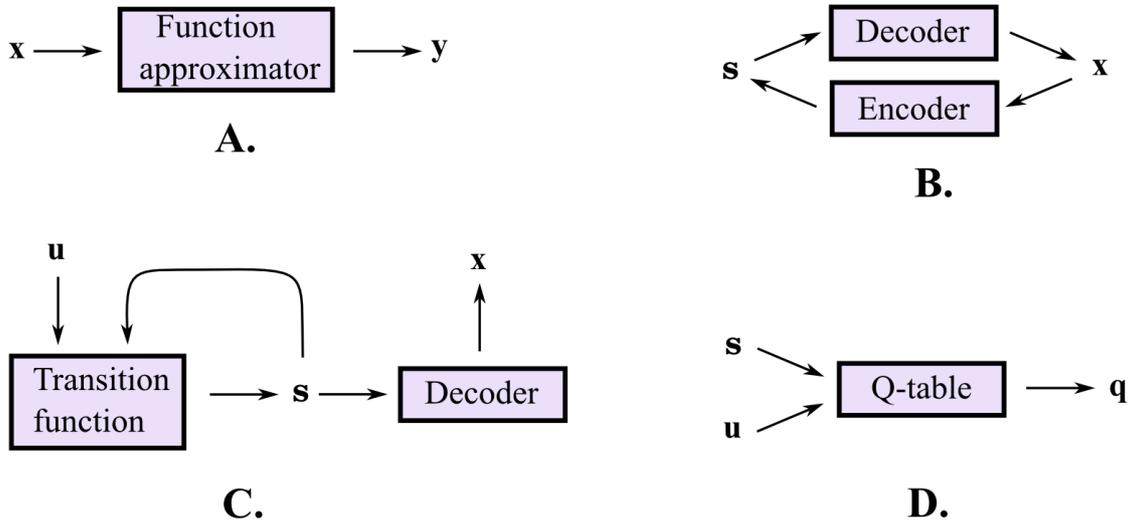

Figure 1. Diagram of some common structures for using neural networks with robots. **A:** Function approximating models are trained to approximate the mappings represented in a training set of pair-wise examples. **B:** Autoencoders can reduce complex or high-dimensional observations to a simple feature representation, often extracting intrinsic information from images. **C:** Recurrent models specialize in dynamics and temporal predictions. **D:** Policy models trained with reinforcement learning seek to plan the best decisions to make under possible future conditions.

### *2.3 Convolutional layers*

Each of the various types of deep learning models are made by stacking multiple layers of regression models. Within these models, different types of layers have evolved for various purposes. One type of layer that warrants particular mention is convolutional layers [48]. Unlike traditional fully connected layers, convolutional layers use the same weights to operate all across the input space. This significantly reduces the total number of weights in the neural network, which is especially important with images that typically have hundreds of thousands to millions of pixels that must be processed.

Processing such images with fully connected layers would require more than $(100K)^2$ to $(1M)^2$ weights connecting each layer, which would be completely impractical. Convolutional layers were inspired by cortical neurons in the visual cortex, which respond only to stimuli with a receptive field. Since convolution approximates this behavior, convolutional layers may be expected to excel at image processing tasks.

The pioneering works in neural networks with convolutional layers (CNNs) applied them to the task of image recognition [48, 49]. Many subsequent efforts built on these, but widespread interest in convolutional layers surged around 2012, when Krizhevsky used them to dominate in the ImageNet image recognition competition [41], and they were able to achieve super-human recognition on other notable image recognition benchmarks and competitions [27, 34, 35]. A flurry of research quickly followed seeking to establish deeper models with improved image processing capabilities [50, 51].

Now, CNNs have become well established as a highly effective deep learning model for a diversity of image-based applications. These applications include semantic image segmentation [52], object localization within images [53], scaling up images with super resolution [54], facial recognition [55, 56], scene recognition [57], and human gesture recognition [58]. Images are not the only type of signal for which CNNs excel. Their capabilities are also effective with any type of signal that exhibits spatio-temporal proximity, such as speech recognition [59], and speech and audio synthesis [60]. Naturally, they have also started to dominate in signal processing domains used heavily in robotics, such as pedestrian detection using LIDAR [61] and micro-Doppler signatures [62], and depth-map estimation [63]. Recent works are even starting to combine signals from multiple modalities and combine them together for unified recognition and understanding [64].

### *2.4 High level trajectory of deep learning with robotics*

Ultimately, the underlying philosophy that prevails in the deep learning community is that every part of a complex system can be made to "learn." Thus, the real power of deep learning does not come from using just one of the structures described in the previous section as a component in a robotics system, but in connecting parts of all of these structures together to form a full system that learns throughout. This is where the "deep" in deep learning begins to make its impact – when each part of a system is capable of learning, the system as a whole can adapt in sophisticated ways.

Neuroscientists are even starting to recognize that many of the patterns evolving within the deep learning community and throughout artificial intelligence are starting to mirror some of those that have previously evolved in the brain [65, 66]. Doya identified that supervised learning methods (Structures A and C) mirror the function of the cerebellum, unsupervised methods (Structure B) learn in a manner comparable to that of the cerebral cortex, and reinforcement learning is analogous with the basal ganglia [67]. Thus, the current trajectory of advancement strongly suggests that control of robots is leading toward full cognitive architectures that divide coordination tasks in a manner increasingly analogous with the brain [68-70].

## 3. Deep learning in robotics

The robotics community has identified numerous goal for robotics in the next 5 to 20 years. These include, but certainly are not limited to, human-like walking and running, teaching by demonstration, mobile navigation in pedestrian environments, collaborative automation, automated bin/shelf picking, automated combat recovery, and automated aircraft inspection and maintenance, and robotic disaster mitigation and recovery [71-75]. This paper identifies seven general challenges for robotics that are critical for reaching these goals and for which DNN technology has high potential for impact:

Challenge 1:  Learning complex, high-dimensional, and novel dynamics.  Analytic derivation of complex dynamics requires human experts, is time consuming, and poses a trade-off between state dimensionality and tractability.  Making such models robust to uncertainty is difficult, and full state information is often unknown.  Systems that can quickly and autonomously adapt to novel dynamics are needed to solve problems such as grasping new objects, traveling over surfaces with unknown or uncertain properties, managing interactions between a new tool and/or environment, or adapting to degradation and/or failure of robot subsystems.  Also needed are methods to accomplish this for systems that possess hundreds (or even thousands) of degrees of freedom, exhibit high levels of uncertainty, and for which only partial state information is available.

Challenge 2:  Learning control policies in dynamic environments.  As with dynamics, control systems that accommodate high degrees of freedom for applications such as multi-arm mobile manipulators, anthropomorphic hands, and swarm robotics are needed.  Such systems will be called upon to function reliably and safely in environments with high uncertainty and limited state information.

Challenge 3:  Advanced manipulation.  Despite advances achieved over 3 decades of active research, robust and general solutions for tasks such as grasping deformable and/or complex geometries, using tools, and actuating systems in the environment (turn a valve handle, open a door, and so forth) remain elusive – especially in novel situations.  This challenge includes kinematic, kinetic, and grasp planning inherent in tasks such as these.

Challenge 4: Advanced object recognition. DNNs have already proven to be highly adept at recognizing and classifying objects [27, 34, 35]. Advanced application examples include recognizing deformable objects and estimating their state and pose for grasping, semantic task and path specification (e.g., go around the table, to the car, and open the trunk), and recognizing the properties of objects and surfaces such as sharp objects that could pose a danger to human collaborators or wet/slippery floors.

Challenge 5: Interpreting and anticipating human actions. This challenge is critical if robots are to work with or amongst people in applications such as collaborative robotics for manufacturing, eldercare, autonomous vehicles operating on public thoroughfares, or navigating pedestrian environments. It will enable teaching by demonstration, which will in turn facilitate task specification by individuals without expertise in robotics or programming. This challenge may also be extended to perceiving human needs and anticipating when robotic intervention is appropriate.

Challenge 6: Sensor fusion & dimensionality reduction. The proliferation of low-cost sensing technologies has been a boon for robotics, providing a plethora of potentially rich, high-dimensional, and multimodal data. This challenge refers to methods for constructing meaningful and useful representations of state from such data.

Challenge 7: High-level task planning. Robots will need to reliably execute high-level commands that fuse the previous six challenges to achieve a new level of utility, especially if they are to benefit the general public. For example, the command "get the milk" must autonomously generate the lower-level tasks of navigating to/from the refrigerator, opening/closing the door, identifying the proper container (milk containers may take many forms), and securely grasping the container.

Loosely speaking, these challenges form a sort of "basis set" for the goals mentioned above. For example, human-like walking and running will rely heavily on Challenges 1 (learning dynamics) and 2 (learning control policies), while teaching by demonstration will require advances in Challenges 4 (object recognition), 5 (interpreting human action), and 6 (sensor fusion).

Table 1 categorizes recent robotics research that utilizes DNN technology according to these challenges, as well as the DNN structures discussed in the previous section. From this several observations are made: First is that Structure A is clearly the most popular DNN architecture in the recent robotics literature. This is likely explained by its intuitive nature, essentially learning to approximate the same function presented to it in the form of training samples. It also requires the least amount of domain knowledge in DNNs to implement. Robotics challenges, however, are not limited to the sort of classification and/or regression problems to which this structure is best suited. Additional focus on applying Structures B, C, and D to robotics problems may very well catalyse significant advancement in many of the identified challenges. One of the purposes of this paper is to emphasize the potential of the other structures to the robotics community.

Somewhat related is the fact that some cells in Table 1 are empty. In the authors' opinion, this is due to a lack of research focus rather than any inherent incompatibilities between challenges and structures. In particular, the ability of Structure B to learn compact representations of state would be particularly useful for estimating the pose, state, and properties of objects (Challenge 4) and the state of human collaborators (Challenge 5).

Table 1. An overview of how DNN structures are used in the recent literature to address the seven challenges.

| | DNN Structure | | | |
| --- | --- | --- | --- | --- |
| | **A** | **B** | **C** | **D** |
| Challenge 1 (Dynamics) | [76, 78, 81, 85, 115, 127] | [87, 112, 113, 115 | [115, 122] | [125, 126] |
| Challenge 2 | [85,115,] | [112] | [122] | [125,128] |
| (Control) | | | | |
| Challenge 3 (Manipulation) | [79, 82-85, 123] | [112] | [123] | [128] |
| Challenge 4 (Object rec.) | [79-81, 88, 123] | | [123] | [128] |
| Challenge 5 (Human actions) | [77, 79, 123, 127] | | [123] | |
| Challenge 6 (Sensor fusion) | [77, 81, 83, 84, 86, 88] | [87, 116, 117] | [114] | [116, 117] |
| Challenge 7 (high-level planning) | | | | [128] |

Table 1 also indicates limited application of DNNs to high-level task planning (Challenge 7). One of the barriers to the application of DNNs is quantifying the quality of such decisions. Standard benchmarks for decision quality are needed. Once this is addressed, DNNs may very well be able to be the tool that allows roboticists to make progress on this very significant challenge.

The balance of this section is categorized by DNN structure and is organized as follows: 1) a discussion of the structure's role in robotics, 2) examples from the recent literature of how the structure is being applied in robotics, and 3) practical recommendations for applying the structure in robotics.

*3.1 Classifiers and discriminative models (Structure A) in robotics*
*3.1.1 The role of Structure A in robotics*

Structure A involves using a deep learning model to approximate a function from sample input-output pairs. This may be the most general-purpose deep learning structure, since there are many different functions in robotics that researchers and practitioners may want to approximate from sample observations. Some examples include mapping from actions to corresponding changes in state, mapping from changes in state to the actions that would cause it, or mapping from forces to motions. Whereas in some cases physical equations for these functions may already be known, there are many other cases where the environment is just too complex for these equations to yield acceptable accuracy. In such situations, learning to approximate the function from sample observations may yield significantly better accuracy.

The functions that are approximated need not be continuous. Function approximating models also excel at classification tasks, such as determining what type of object lies before the robot, which grasping approach or general planning strategy is best suited for current conditions, or what is the state of a certain complex object with which the robot is interacting.

The next section reviews some of the many applications for classifiers, regression models, and discriminative models that have appeared in the recent literature with robotics.

### 3.1.2 Examples in recent research

Punjani and Abbeel [76] used a function approximating deep learning architecture with rectifiers to model the highly coupled dynamics of a radio-controlled helicopter, which is a challenging analytic derivation and difficult system identification problem. Training data was obtained as a human expert flew the helicopter through various aerobatic maneuvers, and the DNN outperformed three state-of-the-art methods for obtaining helicopter dynamics by about 60%.

Neverova et al. [77] modeled how the time between a driver's head movement and the occurrence of a maneuver varies with vehicle speed. The resulting system made predictions every 0.8 seconds based on the preceding 5 seconds of data and anticipated maneuvers about 3.5 seconds before they occurred, with 90.5% accuracy.

A great many works have used function approximating models in the domains of (1) detection and perception, (2) grasping and object manipulation, and (3) scene understanding and sensor fusion. The following three subsections describe recent works in each of these domains.

*Detection and Perception.* DNNs have surged ahead of other models in the domains of detection and perception. They are especially attractive models because they are capable of operating directly on high-dimensional input data instead of requiring feature vectors that are hand-engineered at design time by experts in machine learning and the particular application [1]. This reduces dependence on human experts, and the additional training time may be partially offset by reducing initial engineering effort.

Mariolis, Peleka, and Kargakos [78] studied object and pose recognition for garments hanging from a single point, as if picked by a robotic gripper. Training occurred on pants, shirts, and towels with various size, shape, and material properties, both flat and hanging from various grasp points. On a test set of six objects different from those used in training, the authors achieved 100% recognition and were able to predict grasp point on the garment with a mean error of 5.3 cm. These results were more accurate and faster than support vector machines. Yang, Li, and Fermüller [79] trained a DNN to recognize 48 common kitchen objects and classify human grasps on them from 88 YouTube cooking videos. Notably, the videos were not created with training robots in mind, exhibiting significant variation in background and scenery. Power and precision grasps, and subclasses of each were classified. The system achieved 79% object recognition accuracy and 91% grasp classification accuracy. Chen et al. [80] identified the existence and pose of doors with a convolutional neural network and passed this information to a navigation algorithm for a mobile robot. They suggest that navigating by such visual information can be superior to map-building methods in dynamic environments. Gao, Hendricks, Kuchenbecker, and Darrell [81] integrated both vison- and contact-based perception to classify objects with haptic adjectives (smooth, compliant, etc.). If a robot could predict such information before or quickly after making contact with the objects, it can take appropriate actions such as adjusting its grip on fragile objects or avoiding slippery surfaces. As with the other research studies in this paper, their method did not require manual design of feature vectors from domain-specific knowledge.

*Grasping and object manipulation.* Yu, Weng, Liang, and Xie [82] used a deep convolutional neural network to recognize five known objects resting on a flat surface and categorize their orientation into discretized categories. The study focused on recognition and pose estimation, so grasp planning was limited to positioning a parallel gripper at the object's center and aligned with the estimated angle. Grasping success rates exceeded 90%. Lenz, Lee, and Saxena [83] used deep learning to detect optimal grasps for objects from RGB-D (color + depth) images. The network evaluates numerous potential grasps for the object and identifies the one with the highest potential for success without any prior knowledge of object geometry. Trials on Baxter and PR2 robots resulted in successful grasps 84% and 89% of the time, respectively, compared to 31% for a state-of-the art reference algorithm. Rather than evaluating a set of potential grasps, Redmon and Angelova [84] trained a convolutional neural network to detect an acceptable grasp directly from RGB-D data in one pass. They achieved 88% accuracy and claim real-time performance, arriving at a solution in under 100 milliseconds.

Levine, Pastor, Krizhevsky, and Quillen [85] trained a convolutional neural network to evaluate the potential of a particular robot motion for successfully grasping common office objects from image data, and used a second network to provide continuous feedback through the grasping process. Inspired by hand-eye coordination in humans, the system was robust to object movement and uncertainty in gripper mechanics. While this research may seem similar to visual servoing at first glance, it differs in that no hand-designed feature vectors were required for perception, and transfer functions for closed-loop control were not modeled analytically.

*Scene understanding and sensor fusion*. Extracting meaning from video or still scenes is another application where deep learning has made impressive progress. Neverova, Wolf, Taylor, and Nebout [77] report on their first-place winning entry in the 2014 ChaLearn Looking at People Challenge (http://gesture.chalearn.org), which challenges entrants to recognize 20 different Italian conversational gestures from 13,858 separate RGB-D videos of different people performing those gestures. Ouyang and Wang [86] simultaneously addressed four independent aspects of pedestrian detection with a single DNN: feature extraction, articulation and motion handling, occlusion handling, and classification. They argue that their unified system avoids suboptimal interactions between these usually separate systems. The integrated network was trained on over 60,000 samples from two publically available datasets. Compared to 18 other approaches in the published literature, the authors' system outperforms on both data sets by as much as 9%. Wu, Yildirim, Lim, Freeman, and Tenenbaum [87] attempted to predict the physical outcome of dynamic scenes by vision alone, based on the premise that humans can often predict the outcome of a dynamic scene from visual information – for example, a block sliding down a ramp and impacting another block. They use both simulations from a physics engine and physical trials for training.

Deep learning has also been found to be effective at handling multimodal data generated in robotic sensing applications. Previously mentioned examples include integrating vision and haptic sensor data [81] and incorporating both depth data and image information from RGB-D camera data [77, 83]. Additionally, Schmitz et al. [88] studied tactile object recognition with a *TWENDY-ONE* multi-finger hand, which provides a multimodal set of 312 values from distributed skin sensors, fingertip forces, joint torques, actuator currents, and joint angles. The system was trained on a set of twenty objects – some deliberately similar and some vastly different – handed to the robot in various poses. The investigators achieved an 88% recognition rate, as compared to the 68% using other methods in the literature.

### 3.1.3 Practical recommendations for working with Structure A

Due to their large numbers of meta-parameters, DNNs have developed somewhat of a reputation for being difficult for non-experts to use effectively. However, these parameters also provide significant flexibility, which is a major factor in their overall success. Therefore, training DNNs requires the user to develop at least a basic level of familiarity with several concepts. This section summarizes some of the most important concepts involved in training function approximating DNNs. In particular, applying these techniques will help to address Challenge 4 (advanced object recognition), and to a lesser extent all of the other challenges as well.

Although recent trends lean toward deeper and bigger models, a simple neural network with just one hidden layer and a standard sigmoid-shaped activation function will train much faster, and will provide a useful baseline to give meaning to any improvements from the use of deeper models. When deeper models are used, Leaky Rectifiers tend to promote faster training by diminishing the effects of the vanishing gradient problem [41, 89], and improve accuracy through having a simpler monotonic derivative [91, 91].

Since models with more weights have more flexibility to overfit to the training data, regularization is important role for training the best model. Elastic net combines the well-established $L^1$ and $L^2$ regularization methods to promote robustness against weight saturation and also promote sparsity in the weights [92]. Newer regularization methods, including drop-out [93] and drop-connect [94] have achieved even better empirical results. Several regularization methods also exist specifically to improve robustness with autoencoders [95, 96].

Special-purpose layers can also make a significant difference with DNNs. It is a common practice to alternate between convolutional and max pooling layers. The pooling layers reduce the overall number of weights in the network and also enable the model to learn to recognize objects independent of where they occur in the visual field [97]. Batch normalization layers can yield significant improvements in the rate of convergence by keeping the gradient in a range where it will affect the weights of all neurons [98]. And, residual layers can enable much deeper, and consequently more flexible, models to be trained [99].

To make effective use of deep learning models, it is important to train on one or more General Purpose Graphical Processing Units (GPGPUs) [17]. Many other ways of parallelizing deep neural networks have been attempted, but none of them yet yield the performance gains of GPGPUs [27]. Since DNNs require the use of so many specialized techniques, leveraging an existing toolkit that provides ready-made implementations is an imperative. Fortunately, the deep learning community has been very helpful in releasing open source implementations of new developments, so many well-refined open source deep learning toolkits are now available:

Tensorflow has recently surged in popularity [100]. Theano is a Python-based platform that provides General Purpose Graphical Processing Unit (GPGPU)-parallelization for deep learning [101]. Several popular toolkits build on top of Theano, including as Lasagne and Pylearn2 [102]. Keras is a wrapper around Tensorflow and Theano that seeks to simplify the interfaces for deep learning [103]. Torch offers a Matlab-like environment written in Lua for deep learning, with particular emphasis on convolutional neural networks [104]. In C++, Caffe is one of the more popular toolkits for high-performance convolutional neural networks [105]. It also provides Python bindings. Other C++ toolkits with GPU support are available [106, 107]. Some other toolkits with deep learning support include GroundHog, Theanets [100], Kaldi [109], and CURRENNT [110]. Kustikova gives a survey of many deep learning toolkits for image recognition [111]. Toolkits that employ other parallelization methods, besides GPGPUs, include Hadoop, Mahout, Spark, DeepLearning4j, and Scala.

### 3.2 Generative and Unsupervised models (Structure B) in robotics

#### 3.2.1 The role of Structure B in robotics

One of the characteristic capabilities that make humans so proficient at operating in the real world is their ability to understand what they perceive. A similar capability is offered in autoencoders, a type of deep learning model that both encodes observations into an internal representation, then decodes it back to the original observation. These models digest high-dimensional data and produce compact, low-dimensional internal representations that succinctly describe the meaning in the original observations [3]. Thus, auto-encoders are used primarily in cases where high-dimensional observations are available, but the user wants a low-dimensional representation of state.

Generative models are closely related. They utilize only the decoding portion of an autoencoder, and are useful for predicting observations. Inference methods may be used with generative models to estimate internal representations of state without requiring an encoder to be trained at all. In many ways, generative models may be considered to be the opposite of classifiers, or discriminative models, because they map from a succinct representation to a full high-dimensional set of values similar to those that might typically be observed.

### 3.2.2 Examples in recent research

Finn et al. [112] trained a deep spatial auto-encoder on visual features to extract meaning from the observations and ultimately to achieve visuomotor control. The autoencoder learned how robot actions affected the configuration of objects in the work envelope, and this model was used in a closed-loop controller. Tasks included pushing a block, spooning material into a bowl, scooping with a spatula, and hanging a loop of rope on a hook.

Wu et al. [87] used a generative model to anticipate outcomes from physics simulations. Watter, Springenberg, Bodecker, and Reidmiller [113] also applied a generative model to model the nonlinear dynamics of simple physical systems and control them. Noda, Arie, Suga, and Ogata [114] developed a novel deep learning solution involving both unsupervised methods and recurrent models for integrating multi-modal sensorimotor data, including RGB images, sound data, and joint angles.

Another example of Structure B was demonstrated by Polydoros, Nalpantidis, and Kruger in modeling the inverse dynamics of a manipulator [115]. The network was trained using state variables recorded while operating under standard closed-loop control. A "fading memory" feature allowed the DNN to adapt as dynamics changed with payload and mechanical wear. Analytic dynamic models have difficulty coping with such changes, and are difficult to derive for highly compliant serial-elastic manipulators such as the new class of collaborative robots. The authors report that their system outperforms the state-of-the-art in real-time learning evaluations and converges quickly, even with noisy sensor data.

Günther, Pilarski, Helfrich, Shen, and Diepold [116, 117] designed a DNN to automatically create meaningful feature vectors. The network was able to extract low-dimensional features from high-dimensional camera images of welds in a laser welding process. These features were subsequently used with other machine learning and control strategies to close the loop on the welding process.

### 3.2.3 Practical recommendations for working with Structure B

Autoencoders and other unsupervised DNN techniques are particularly well suited for addressing challenges pertaining to high-dimensional observations (1 and 6). They both reduce dimensionality and extract meaningful representations of state, which is the first step in effective sensor fusion.

Convolutional layers are well known to be effective for digesting images. Since images are common with robots, autoencoders that use convolutional layers in their encoding portion tend to be especially effective for estimating state from images [118]. For the decoding portion of the autoencoder, convolutional layers offer little advantage. A somewhat less-known technique involves training the decoder to predict only a single pixel and parameterizing the decoder to enable the user to specify which pixel it should predict [119]. This approach has many analogies with convolution, and experimentally seems to lead to much faster training times.

Regularization is particularly important for achieving good results with autoencoders, and specialized regularization methods have been designed particularly for autoencoders [120]. However, some experiments have shown that instead of heavily regularizing the encoder, it may even work better to entirely omit the encoder, and just use a standalone decoder [119]. In this configuration, the internal representation of state is inferred in a latent manner by using gradient descent until the internal representation converges with the decoder. Nonlinear dimensionality reduction methods have also been shown to be effective for pretraining such latent representations [119].

### 3.3 Recurrent models (Structure C) in robotics

#### 3.3.1 The role of Structure C in robotics

Recurrent models excel at learning to anticipate complex dynamics. The recurrent connections in such models give them a form of "memory" that they can use to remember the current state. This knowledge of state enables them to model the effects of time in a changing environment.

#### 3.3.2 Examples in recent research

Jain et al. [121] trained a recurrent architecture to predict traffic maneuvers in a human-driven automobile in an effort to improve current collision avoidance systems which often do not intervene in time to avoid an accident. Multimodal data inputs included video of the driver, video of the road in front of the car, dynamic state of the vehicle, GPS coordinates, and street maps of the area around the car.

Several researchers have used recurrent networks to deduce system dynamics directly from full observations. Lenz, Knepper, and Saxena [122] modeled robotic food cutting with a knife. This includes difficult-to-model effects such as friction, deformation, and hysteresis. Food-knife surface contact changes through the cut, and so do the material properties of the food, as when passing between the peel and the center of a fruit. Data obtained while operating under fixed-trajectory stiffness control was used to train the DNN on the system dynamics, and the resulting model was used to implement a model predictive control algorithm (a variation of Structure D). Their system outperformed fixed-trajectory stiffness control, increasing mean cutting rate from 1.5 cm/s to 5.1 cm/s.

Hwang et al. [123] demonstrated gesture recognition with a recurrent model, and coordinated it with attention switching, object perception, and grasping. The robot focused on a human collaborator, who gestured to one of two objects. The robot then switched its focus to the indicated object, recognized the object, and found an acceptable grasp. Their system achieved a successful grasp 85% of the time when simulated on an iCub humanoid robot.

### 3.3.3 Practical recommendations for working with Structure C

Recurrent models are well suited for addressing challenges pertaining to the complexities of temporal effects (Challenges 1, 5, and 7). This section describes some recommendations for working effectively with recurrent models.

Unfortunately, recurrent models have a somewhat negative reputation for being difficult to train. One of the biggest problems is that training them with gradient methods requires unfolding through time, which effectively makes them behave as networks that are much deeper than they already are. Given so much depth, the training gradients tend to become vanishingly small [10, 11]. This problem was largely solved by the error-carousel idea in LSTM networks [124], so it would be helpful to become familiar with that solution before attempting to work with recurrent models.

When observations are very high dimensional, such as occurs when digital images are used with robots, a much simpler solution becomes possible. This solution is to simply infer the intrinsic state from the images. If the state can be inferred accurately, then the recurrent essentially goes away, making it possible to train the structure with example pairs presented in arbitrary order, just like Structure A [22]. Even if the internal state can only be inferred with a small degree of accuracy, this still provides useful pre-training, which may significantly reduce the necessary training time with a recurrent model [119].

### 3.4 Policy learning models (Structure D) in robotics

#### 3.4.1 The role of Structure D in robotics

Learning a near optimal (or at least a reasonably acceptable) control policy is often the primary objective in combining machine learning with robotics. The canonical model for using deep neural networks for learning a control policy is deep Q-learning [47]. It uses a DNN to model a table of Q-values, which are trained to converge to a representation of the values for performing each possible action in any state. Although Structure D is quite similar to Structure A in terms of the model itself, they are trained in significantly different ways. Instead of minimizing prediction error against a training set of samples, deep Q-networks seek to maximize long-term reward. This is done through seeking a balance between exploration and exploitation that ultimately leads to an effective policy model.

Ultimately, reinforcement learning models are useful for learning to operate dynamic systems from partial state information, and controllers based on deep reinforcement learning can be very computationally efficient at runtime [125]. They automatically infer priorities based on rewards that are obtained during training. In theory, they provide a complete control policy learning system, but they do suffer from extremely slow training times. Consequently, many of the works in the next section combine them with other approaches in order to seek greater levels of control accuracy and training speed.

*3.4.2 Examples in recent research*

Zhang, Kahn, and Levine [125] learned a control policy to implement a model predictive control guided search for autonomous aerial vehicles. Reducing computational load in mobile robotics translates into power savings that increase range and/or improve performance. Without the need for full state information, fewer onboard sensors are required, further reducing power consumption, cost, and weight.

Deep reinforcement learning has also been used to control dynamic systems from video, without direct access to state information. Lillicrap, Hunt, and Pritzel [126] trained a deep reinforcement learner based on pixel data over 20 simulated dynamic systems and developed a motion planning system that performed as well or better than algorithms that take advantage of the full state of the dynamic system.

Finn, Levine, and Abbeel [127] used video of human experts performing various tasks to train a DNN to learn nonlinear cost functions (with Structure A). Once these cost functions were learned, they could be used to train a reinforcement learner (Structure D) for motion planning. They demonstrated the ability to complete tasks that involved complex 2nd-order dynamics and hard-to-model interactions between a manipulator and various objects, including 2D navigation, reaching, peg insertion, placing a dish, and pouring.

Visuomotor control requires an even closer integration between object perception and grasping, mapping image data directly to actuator control signals. Levine, Finn, Darrell, and Abbeel [128] used reinforcement learning (Structure D) to show that this can be superior to separate systems for perception and control. Test applications included shape sorting, screwing a cap onto a bottle, fitting a hammer claw to a nail, and placing a coat hanger on a rack. The resulting system could perform the tasks reliably, even with moderate visual distractors.

As mentioned earlier, Günther, Pilarski, Helfrich, Shen, and Diepold [116, 117] combined autoencoders with reinforcement learning models to control a laser welding system from camera images.

### 3.4.3 Practical recommendations for working with Structure D

Policy learning models are ultimately the solution to addressing Challenges 2 (learning control policies in dynamic environments) and 7 (high-level task planning). Perhaps, the biggest difficulty when working with reinforcement learning models, however, is the huge amount of computation time necessary to train them. Although such models are highly efficient after training, they tend to require significantly more training pattern presentations before they converge to represent reliable control policies. Taking care to find an efficient GPU-optimized implementation, therefore, can make a big difference. Another important technique is to train in simulation before attempting to train with an actual robot. This reduces wear on physical equipment, as well reduces training time. Even if only a crude simulation is available, a model that has been pre-trained on a similar challenge will converge much more quickly to fit the real challenge than one that was trained from scratch.

Since robots often operate in a space with continuous actions, traditional Q-learning is not directly applicable. Actor-critic models, however, address this problem nicely. They regress actions in conjunction with the continuous Q-table, and lead to a final model that directly computes the best action given the current observation, which is well suited for robotics applications [129].

Another important consideration is the exploration policy. The traditional epsilon-greedy exploration policy leads to much higher computational training requirements than modern approaches [130, 131]. It is, therefore, advantageous to train Structure D in an approach that intelligently explores novel states.

## 4. Current Shortcomings of DNNs for Robotics

For all of its benefits, deep learning does pose some drawbacks. Perhaps most significant is the volume of training data required, which is particularly problematic in robotics because generating training data on physical systems can be expensive and time consuming. For instance, Levine et al. [85] used 14 robots to collect over 800,000 grasp attempts over a period of 2 months. Jain et al. [121] trained their traffic maneuver prediction system on 1180 miles of high- and low-speed driving with 10 different drivers. Punjani and Abbeel [76] required repeated demonstrations of helicopter aerobatic maneuvers by a human expert. Neverova et al. [77] had access to over 13,000 videos of conversations, and Ouyang and Wang [86] had access to 60,000 samples for pedestrian detection. Pinto and Gupta [132] needed 700 hours of robot time to generate a data set of 50,000 grasps for the training of a convolutional neural network.

Despite this, the literature does contain clever approaches to mitigating this disadvantage. One approach entails using simulation to generate virtual training data. For example, Mariolis et al. 78] pre-trained their garment pose recognition networks on a large synthetic data set created in simulation using 3D graphics software. Kappler, Bohg, and Schaal [133] generated a database of over 300,000 grasps on over 700 objects in simulation, generating physics-based grasp quality metrics for each and using this to classify grasp stability automatically. They validated via human classification of grasps and concluded that the computer- and human-generated labeling had good correlation. Another strategy is leveraging training data through digital manipulation. Neverova et al. [77] faced the challenge that speed of conversational gestures varies significantly among different people. They varied video playback speed to simulate this temporal variance, expanding their training set without the need to acquire additional samples. Still other researchers utilizing reinforcement learning, such as Polydoros et al. [115] and Zhang et al. [125], automated training using alternative control systems during the learning phase.

Training time is another challenge associated with the sheer size of DNNs. Typical models involve up to millions of parameters and can take days to train on parallel hardware, which is practical only for frequently repeated tasks that provide adequate payback on training time invested. One way to reduce training time is distributing a task among multiple, smaller DNNs. Mariolis et al. [78] trained two DNNs: One performed object classification, and its result was passed to a second network for pose recognition. This multi-step approach sped both training and classification at runtime. Lenz et al. [122] employed a two-stage network design for grasp detection. The first DNN had relatively few parameters. Sacrificing accuracy for speed, it eliminated highly unlikely grasps. The second stage had more parameters, making it more accurate, but was relatively quick since it did not need to consider unlikely grasps. They found the combination to be robust and computationally efficient. It should be noted, however, that this strategy represents a tradeoff with other researchers' suggestions that integrating multiple functions within a single network results in better performance [86, 128].

The work of Zhang et al. [125] highlights two additional challenges. First, unsupervised learning is not practical for robotic systems where a single failure is catastrophic, as in aerial vehicles. Second, providing the necessary computational resources for deep learning in a system that is sensitive to weight, power consumption, and cost is often not practical. The authors trained their aerial systems using a ground-based control system communicating wirelessly with the vehicle. This made training safe and automatic, and allowed them to use off-board computing resources for training.

## 5. Conclusion

Deep learning has shown promise in significant sensing, cognition, and action problems, and even the potential to combine these normally separate functions into a single system. DNNs can operate on raw sensor data and deduce key features in that data without human assistance, potentially greatly reducing up-front engineering time. They are also adept at fusing high-dimensional, multimodal data. Improvement with experience has been demonstrated, facilitating adaptation in the dynamic, unstructured environments in which robots operate.

Some remaining barriers to the adoption of deep learning in robotics include the necessity for large training data and long training times. Generating training data on physical systems can be relatively time consuming and expensive. One promising trend is crowdsourcing training data via cloud robotics [134]. It is not even necessary that this data be from other robots, as shown by Yang's use of general-purpose cooking videos for object and grasp recognition [79]. Regarding training time, local parallel processing [17] and increases in raw processing speed have led to significant improvements. Distributed computing offers the potential to direct more computing resources to a given problem [88] but can be limited by communication speeds [2]. There may also be algorithmic ways of making the training process more efficient yet to be discovered. For example, deep learning researchers are actively working on directing the network's attention to the most relevant subspaces within the data and applying biologically inspired, sparse DNNs with fewer synaptic connections to train [27].

Ultimately, the trends are moving toward greater levels of cognition, and some researchers even believe that deep learning may achieve human-level abilities in the near future [1, 134]. However, deep learning still has many obstacles to overcome before achieving such an ambitious objective. Currently, cognitive training datasets do not even exist [134]. Although DNNs excel at 2D image recognition, they are known to be highly susceptible to adversarial samples [135], and they still struggle to model 3D spatial layouts with object invariance [65]. Currently, DNNs appear to be powerful tools for practitioners in robotics, but only time will tell whether they can really deliver the capabilities that are needed for dexterous adaptation in general environments.

## References

[1] LeCun Y, Bengio Y, Hinton G. Deep learning. Nature. 2015;521(7553):436-444.

[2] Jordan MI, Mitchell TM. Machine learning: arends, perspectives, and prospects. Science. 2015;349(6245):255-260.

[3] Böhmer W, Springenberg JT, Boedecker J, et al. Autonomous learning of state representations for control: an emerging field aims to autonomously learn state representations for reinforcement learning agents from their real-world sensor observations. KI-Künstliche Intelligenz. 2015;29(4):353-362.

[4] Stigler SM. Gauss and the invention of least squares. Ann of Statistics. 1981;9(3):465-474.

[5] Haykin S. Neural networks: a comprehensive foundation. 2nd ed. Upper Saddle River, New Jersey: Prentice Hall; 2004.

[6] Bryson AE, Denham WF, Dreyfus SE. Optimal programming problems with inequality constraints. AIAA Journal. 1963;1(11):2544-2550.

[7] Rumelhart DE, Hinton GE, Williams RJ. Learning representations by back-propagating errors. Nature. 1986;323:533-536.

[8] Werbos P. Beyond regression: new tools for prediction and analysis in the behavioral sciences. [Ph.D. dissertation]. Dept. Statistics, Harvard Univ.; 1974.

[9] Cybenko G. Approximation by superpositions of a sigmoidal function. Math of Control, Signals and Sys. 1989;2(4):303-314.

[10] Hochreiter S. Untersuchungen zu dynamischen neuronalen netzen. [Master's thesis]. Institut Fur Informatik, Technische Universitat; 1991.

[11] Hochreiter S. The vanishing gradient problem during learning recurrent neural nets and problem solutions. Int. J. of Uncertainty, Fuzziness and Knowledge-Based Syst. 1998;6(2).

[12] Miyamoto H, Kawato M, Setoyama T, & Suzuki R. Feedback-error-learning neural network for trajectory control of a robotic manipulator. Neural Networks 1998;1(3):251-265.

[13] Lewis FW, Jagannathan S, & Yesildirak A. (1998). Neural network control of robot manipulators and non-linear systems. CRC Press.

[14] Miller WT, Werbos PJ, & Sutton RS. (1995). Neural networks for control. MIT Press.

[15] Lin CT., & Lee CSG. Neural-network-based fuzzy logic control and decision system. IEEE Transactions on Computers. 1991;40(12):1320-1336.

[16] Pomerleau DA. (1989). ALVINN, an autonomous land vehicle in a neural network (No. AIP-77). Carnegie Mellon University, Computer Science Department.

[17] Oh K, Jung K. GPU implementation of neural networks. Pattern Recognition. 2004;37(6):1311-1314.

[18] Hinton GE, Osindero S, Teh Y. A fast learning algorithm for deep belief nets. Neural Computation. 2006;18(7):1527-1554.

[19] Dean J, Corrado G, Monga R, et al. Large scale distributed deep networks. Advances in Neural Information Process. Syst. 25; 2012.

[20] Tani J, Ito M, & Sugita Y. Self-organization of distributedly represented multiple behavior schemata in a mirror system: reviews of robot experiments using RNNPB. Neural Networks. 2004;17(8):1273-1289.

[21] Ijspeert AJ. Central pattern generators for locomotion control in animals and robots: a review. Neural Networks. 2008;21(4):642-653.

[22] Gashler M, Martinez T. Temporal nonlinear dimensionality reduction. Neural Networks (IJCNN), 2011 International Joint Conference on; 2011. p. 1959-1966.

[23] Pomerleau DA (2012). Neural network perception for mobile robot guidance (Vol. 239). Springer Science & Business Media.

[24] Thrun S. Learning to play the game of chess. Advances in Neural Inform. Process. Syst.: Proc. of the 1994 Conf.

[25] Campbell M, Hoane AJ, Hsu F. Deep blue. Artificial Intelligence. 2002;134(1):57-83.

[26] Pinto N, Cox DD, DiCarlo JJ. Why is real-world visual object recognition hard? PLoS Computational Biology. 2008;4(1).

[27] Schmidhuber J. Deep learning in neural networks: an overview. Neural Networks. 2015;6(1)85-117.

[28] Graves A, Liwicki M, Fernández S, et al. A novel connectionist system for unconstrained handwriting recognition. Pattern Anal and Machine Intell., IEEE trans. on. 2009;31(5):855-868.

[29] Yang M, Ji S, Xu W, et al. Detecting human actions in surveillance videos. TREC Video Retrieval Evaluation Workshop; 2009.

[30] Lin M, Chen Q, Yan S. Network in network. 2013. Available: https://arxiv.org/abs/1312.4400

[31] Ciresan D, Giusti A, Gambardella LM, et al. Deep neural networks segment neuronal membranes in electron microscopy images. Advances in Neural Information Processing Sys 25; 2012.

[32] Roux L, Racoceanu D, Lomenie N, et al. Mitosis detection in breast cancer histological images an ICPR 2012 contest. J Pathol Inform. 2013;4(8).

[33] Cireşan DC, Giusti A, Gambardella LM, et al. Mitosis detection in breast cancer histology images with deep neural networks. In: K. Mori, I. Sakuma, Y. Sato, C. Barillot and N. Navab, editors. Medical Image Computing and Computer-Assisted Intervention–MICCAI 2013. Springer; 2013.

[34] Cireşan D, Meier U, Masci J, et al. A committee of neural networks for traffic sign classification. Neural Networks (IJCNN), 2011 Int. Joint Conf. on; 2011. p. 1918-1921.

[35] Ciresan D, Meier U, Schmidhuber J. Multi-column deep neural networks for image classification. Computer Vision and Pattern Recognition (CVPR), 2012 IEEE Conference on; 2012. p. 3642-3649.

[36] Dunne RA, & Campbell NA. On the pairing of the softmax activation and cross-entropy penalty functions and the derivation of the softmax activation function. In Proc. 8th Aust. Conf. on the Neural Networks, Melbourne 1997;181 (Vol. 185).

[37] Wilson DR, Martinez TR. The general inefficiency of batch training for gradient descent learning. Neural Networks. 2003;16(10):1429-1451.

[38] Tieleman T, Hinton G. (2012). Lecture 6.5-rmsprop: divide the gradient by a running average of its recent magnitude. COURSERA: Neural Networks for Machine Learning, 4(2).

[39] Kingma D, Ba J. (2014). Adam: a method for stochastic optimization. arXiv preprint arXiv:1412.6980.

[40] Vincent P, Larochelle H, Lajoie I, et al. Stacked denoising dutoencoders: learning useful representations in a deep network with a local denoising criterion. J Mach Learning Research. 2010;11:3371-3408.

[41] Krizhevsky A, Sutskever I, Hinton GE. Imagenet classification with deep convolutional neural networks. Advances in Neural Information Processing Systems 25; 2012.

[42] LeCun Y, Bengio Y. Convolutional networks for images, speech, and time series. In M. Arbib, editor. The Handbook of Brain Theory and Neural Networks. 2nd edition. Cambridge, MA: MIT Press; 2003.

[43] Werbos PJ. Backpropagation through time: what it does and how to do it. Proc. IEEE. 1990;78(10):1550-1560.

[44] Sjöberg J, Zhang Q, Ljung L, et al. Nonlinear black-box modeling in system identification: a unified overview. Automatica. 1995;31(12):1691-1724.

[45] Hochreiter S, Schmidhuber J. Long short-term memory. Neural Computation. 1997;9(8):1735-1780.

[46] Atkeson CG, Santamaria JC. A comparison of direct and model-based reinforcement learning. Robotics and Automation, IEEE Int. Conf. on; Albuquerque, NM. 1997. p. 3557-3564.

[47] Mnih V, Kavukcuoglu K, Silver D, et al. Human-level control through deep reinforcement learning. Nature. 2015;518:529-533.

[48] LeCun Y, Boser B, Denker JS, Henderson D, Howard RE, Hubbard W, Jackel LD. Backpropagation applied to handwritten zip code recognition. Neural Computation. 1989;1(4), 541-551.

[49] LeCun Y, Bottou L, Bengio Y, Haffner P.. Gradient-based learning applied to document recognition. Proceedings of the IEEE. 1998;86(11):2278-2324.

[50] Simonyan K, Zisserman A. (2014). Very deep convolutional networks for large-scale image recognition. arXiv preprint arXiv:1409.1556.

[51] Szegedy C, Liu W, Jia Y, Sermanet P, Reed S, Anguelov D., ... Rabinovich A. (2015). Going deeper with convolutions. In Proc of the IEEE Conference on Computer Vision and Pattern Recognition (pp. 1-9).

[52] Chen LC, Papandreou G, Kokkinos I, Murphy K, & Yuille AL. (2014). Semantic image segmentation with deep convolutional nets and fully connected crfs. arXiv preprint arXiv:1412.7062.

[53] Sermanet P, Eigen D, Zhang X, Mathieu M, Fergus R, LeCun Y. (2013). Overfeat: integrated recognition, localization and detection using convolutional networks. arXiv preprint arXiv:1312.6229.

[54] Dong C, Loy CC, He K, Tang X. (2014, September). Learning a deep convolutional network for image super-resolution. In European Conf on Computer Vision (pp. 184-199). Springer International Publishing.

[55] Sun Y, Wang X, Tang X. (2013). Deep convolutional network cascade for facial point detection. In Proc of the IEEE Conf on Computer Vision and Pattern Recognition (pp. 3476-3483).

[56] Taigman Y, Yang M, Ranzato MA, Wolf L. (2014). Deepface: closing the gap to human-level performance in face verification. In Proc of the IEEE Conf on Computer Vision and Pattern Recognition (pp. 1701-1708).

[57] Zhou B, Lapedriza A, Xiao J, Torralba A, Oliva A. (2014). Learning deep features for scene recognition using places database. In Advances in Neural Information Processing Sys (pp. 487-495).

[58] Ji S, Xu W, Yang M, Yu K.. 3D convolutional neural networks for human action recognition. IEEE Transactions on Pattern Analysis and Machine Intelligence. 2013;35(1):221-231.

[59] Graves A, Mohamed AR, Hinton G. (2013, May). Speech recognition with deep recurrent neural networks. In Acoustics, Speech and Signal Proc (ICASSP), 2013 IEEE Int Conf on (pp. 6645-6649). IEEE.

[60] Bakshi BR, Stephanopoulos G. Wave-net: a multiresolution, hierarchical neural network with localized learning. AIChE Journal. 1993;39(1):57-81.

[61] Arel I, Rose DC, Karnowski TP. Deep machine learning - a new frontier in artificial intelligence research [research frontier]. IEEE Computational Intelligence Magazine. 2010;5(4);13-18.

[62] Kim Y, Moon T. . Human detection and activity classification based on micro-doppler signatures using deep convolutional neural networks. IEEE Geoscience and Remote Sensing Letters. 2016;13(1):8-12.

[63] Liu F, Shen C, Lin G, Reid I.. Learning depth from single monocular images using deep convolutional neural fields. IEEE Transactions on Pattern Analysis and Machine Intelligence 2016;38(10):2024-2039.

[64] Eitel A, Springenberg JT, Spinello L, Riedmiller M, Burgard W. (2015, September). Multimodal deep learning for robust rgb-d object recognition. In Intelligent Robots and Sys (IROS), 2015 IEEE/RSJ International Conference on (pp. 681-687). IEEE.

[65] Bohannon J. Helping robots see the big picture. Science. 2014;346(6206):186-187.

[66] Sweller J. (2008, September). Evolutionary bases of human cognitive architecture: implications for computing education. In Proc of the Fourth Int Workshop on Computing Education Research (pp. 1-2). ACM.

[67] Doya K. What are the computations of the cerebellum, the basal ganglia and the cerebral cortex? Neural Networks, 1999;12(7):961-974.

[68] Langley P, Laird JE, Rogers S. Cognitive architectures: research issues and challenges. Cognitive Sys Research 2009;10(2):141-160.

[69] Sun R. (2006). Cognition and multi-agent interaction: from cognitive modeling to social simulation. Cambridge University Press.

[70] Duch W, Oentaryo RJ, Pasquier M. (2008, June). Cognitive Architectures: Where do we go from here? In AGI (Vol. 171, pp. 122-136).

[71] A roadmap for US robotics: from internet to robotics, 2016 edition.

[72] World Technology Evaluation Center, Inc. International Assessment of Research and Development in Robotics. Baltimore, MD, USA; 2006.

[73] FY2009-2034 Unmanned systems integrated roadmap. Washington, DC: Department of Defence (US); 2009.

[74] Material Handling Institute. Material handling and logistics U.S. roadmap 2.0. 2017.

[75] DARPA Robotics Challenge [Internet]. [cited 2017 May 20]. Available from: http://www.darpa.mil/program/darpa-robotics-challenge

[76] Punjani AP, Abbeel P. Deep learning helicopter dynamics models. Robotics and Automation (ICRA), 2015 IEEE International Conference on; 2015. p. 3223-3230.

[77] Neverova N, Wolf C, Taylor GW, et al. Multi-scale deep learning for gesture detection and localization. Computer Vision-ECCV 2014 Workshops; 2014. p. 474-490.

[78] Mariolis I, Peleka G, Kargakos A, et al. Pose and category recognition of highly deformable objects using deep learning. Advanced Robotics (ICAR), 2015 International Conference on; Istanbul. 2015. p. 655-662.

[79] Yang Y, Li Y, Fermüller C, et al. Robot learning manipulation action plans by watching unconstrained videos from the world wide web. 29th AAAI Conference on Artificial Intelligence (AAAI-15); Austin, TX. 2015.

[80] Chen W, Qu T, Zhou Y, et al. Door recognition and deep learning algorithm for visual based robot navigation. Robotics and Biomimetics (ROBIO), 2014 IEEE International Conference on; Bali. 2014. p. 1793-1798.

[81] Gao Y, Hendricks LA, Kuchenbecker KJ, et al. Deep learning for tactile understanding from visual and haptic data. 2015. Available: http://arxiv.org/abs/1511.06065

[82] Yu J, Weng K, Liang G, et al. A vision-based robotic grasping system using deep learning for 3D object recognition and pose estimation. Robotics and Biomimetics (ROBIO), 2013 IEEE International Conference on; Shenzhen. 2013. p. 1175-1180.

[83] Lenz I, Lee H, Saxena A. Deep learning for detecting robotic grasps. Int J Robotics res. 2015;34(4-5):705-724.

[84] Redmon J, Angelova A. Real-time grasp detection using convolutional neural networks. 2015 IEEE International Conference on Robotics and Automation (ICRA); Seattle, WA. 2015. p. 1316-1322.

[85] Levine S, Pastor P, Krizhevsky A, et al. Learning hand-eye coordination for robotic grasping with deep learning and large-scale data collection. 2016. Available: http://arxiv.org/abs/1603.02199

[86] Ouyang W, Wang X. Joint deep learning for pedestrian detection. Computer Vision, 2013 IEEE Int. Conf. on; Sidney, VIC. 2013. p. 2056-2063.

[87] Wu J, Yildirim I, Lim JJ, et al. Galileo: Perceiving physical object properties by integrating a physics engine with deep learning. Advances in Neural Information Processing Systems 28; 2015.

[88] Schmitz A, Bansho Y, Noda K, et al. Tactile object recognition using deep learning and dropout. Humanoid Robots, 2014 14th IEEE-RAS Int. Conf. on; 2014. p. 1044-1050.

[89] Zheng H, Yang Z, Liu , Liang J, Li Y. (2015, July). Improving deep neural networks using softplus units. In Neural Networks (IJCNN), 2015 Int Joint Conf on (pp. 1-4). IEEE.

[90] Xu B, Wang N, Chen T, Li M. (2015). Empirical evaluation of rectified activations in convolutional network. arXiv preprint arXiv:1505.00853.

[91] Gashler MS, Ashmore SC. (2016). Modeling time series data with deep Fourier neural networks. Neurocomputing, 188, 3-11.

[92] Zou H, Hastie T. Regularization and variable selection via the elastic net. J Royal Statistical Society: Series B (Statistical Methodology), 2005;67(2):301-320.

[93] Srivastava N, Hinton GE, Krizhevsky A, Sutskever I, Salakhutdinov R. Dropout: a simple way to prevent neural networks from overfitting. Journal of Machine Learning Research, 2014;15(1):1929-1958.

[94] Wan L, Zeiler M, Zhang S, Cun YL, Fergus R. (2013). Regularization of neural networks using dropconnect. In Proc of the 30th Int Conf on Machine Learning (ICML-13) (pp. 1058-1066).

[95] Rifai S, Vincent P, Muller X, Glorot X, Bengio Y. (2011). Contractive auto-encoders: explicit invariance during feature extraction. In Proc of the 28th Int Conf on Machine Learning (ICML-11) (pp. 833-840).

[96] Li Z, Fan Y, Liu W. The effect of whitening transformation on pooling operations in convolutional autoencoders. EURASIP J on Advances in Signal Proc, 2015; (1):37.

[97] Jarrett K, Kavukcuoglu K, LeCun Y. (2009, September). What is the best multi-stage architecture for object recognition? In Computer Vision, 2009 IEEE 12th Int Conf on (pp. 2146-2153). IEEE.

[98] Ioffe S, Szegedy C. (2015). Batch normalization: accelerating deep network training by reducing internal covariate shift. arXiv preprint arXiv:1502.03167.

[99] He K, Zhang X, Ren S, Sun J. (2016). Deep residual learning for image recognition. In Proc of the IEEE Conf on Computer Vision and Pattern Recognition (pp. 770-778).

[100] Abadi M, Agarwal A, Barham P, Brevdo E, Chen Z, Citro C, ... Ghemawat S. (2016). Tensorflow: large-scale machine learning on heterogeneous distributed systems. arXiv preprint arXiv:1603.04467.

[101] Bergstra J, Bastien F, Breuleux O, Lamblin P, Pascanu R, Delalleau O, ... Bengio Y. (2011). Theano: deep learning on gpus with python. In NIPS 2011, BigLearning Workshop, Granada, Spain (Vol. 3).

[102] Goodfellow IJ, Warde-Farley D, Lamblin P, Dumoulin V, Mirza M, Pascanu R,... Bengio,Y. (2013). Pylearn2: a machine learning research library. arXiv preprint arXiv:1308.4214.

[103] Chollet F. (2015). Keras: deep learning library for theano and tensorflow. URL: https://keras. io/k.

[104] Collobert R, Kavukcuoglu K, Farabet C. (2011). Torch7: a matlab-like environment for machine learning. In BigLearn, NIPS Workshop (No. EPFL-CONF-192376).

[105] Jia Y, Shelhamer E, Donahue J, Karayev S, Long J, Girshick R, ... Darrell T. (2014, November). Caffe: convolutional architecture for fast feature embedding. In Proc 22nd ACM Int Conf on Multimedia (pp. 675-678). ACM.

[106] Attardi G. (2015, June). Deepnl: a deep learning nlp pipeline. In Proc of NAACL-HLT (pp. 109-115).

[107] Gashler M. Waffles: a machine learning toolkit. J Machine Learning Res, 2011;12(Jul):2383-2387.

[108] Johnson L. Theanets documentation. Url: http://theanets.readthedocs.org/en/stable/

[109] Povey D, Ghoshal A, Boulianne G, Burget L, Glembek O, Goel N,... Silovsky J. (2011). The Kaldi speech recognition toolkit. In IEEE 2011 workshop on automatic speech recognition and understanding (No. EPFL-CONF-192584). IEEE Signal Processing Society.

[110] Weninger F, Bergmann J, Schuller BW. Introducing CURRENNT: the munich open-source CUDA recurrent neural network toolkit. J Machine Learning Research, 2015;16(3):547-551.

[111] Kustikova VD, Druzhkov PN. (2014). A survey of deep learning methods and software for image classification and object detection. OGRW2014, 5.

[112] Finn C, Tan XY, Duan Y, et al. Deep spatial autoencoders for visuomotor learning. 2015. Available: http://arxiv.org/abs/1509.06113

[113] Watter M, Springenberg J, Boedecker J, et al. Embed to control: a locally linear latent dynamics model for control from raw images. Advances in Neural Information Proc Sys 28; 2015.

[114] Noda K, Arie H, Suga Y, et al. Multimodal integration learning of robot behavior using deep neural networks. Robotics and Autonomous Sys. 2014;62(6):721-736.

[115] Polydoros AS, Nalpantidis L, Kruger V. Real-time deep learning of robotic manipulator inverse dynamics. Intelligent Robots and Systems (IROS), 2015 IEEE/RSJ International Conference on; 2015. p. 3442-3448.

[116] Günther J, Pilarski PM, Helfrich G, et al. Intelligent laser welding through representation, prediction, and control learning: an architecture with deep neural networks and reinforcement learning. Mechatronics. 2015;34:1-11.

[117] Günther J, Pilarski PM, Helfrich G, et al. First steps towards an intelligent laser welding architecture using deep neural networks and reinforcement learning. Procedia Technology. 2014;15:474-483.

[118] Masci J, Meier U, Cireşan D, Schmidhuber J. (2011). Stacked convolutional auto-encoders for hierarchical feature extraction. Artificial Neural Networks and Machine Learning–ICANN 2011, 52-59.

[119] Ashmore SC, Gashler MS. (2016, December). Practical techniques for using neural networks to estimate state from images. In Machine Learning and Applications (ICMLA), 2016 15th IEEE Int Conf on (pp. 916-919). IEEE.

[120] Rifai S, Vincent P, Muller X, Glorot X, Bengio Y. (2011). Contractive auto-encoders: explicit invariance during feature extraction. In Proc of the 28th Int Conf on Machine Learning (ICML-11) (pp. 833-840).

[121] Jain A, Koppula HS, Soh S, et al. Brain4Cars: car that knows before you do via sensory-fusion deep learning architecture. 2016. Available: http://arxiv.org/abs/1601.00740

[122] Lenz I, Knepper R, Saxena A. Deepmpc: learning deep latent features for model predictive control. Robotics: Science and Systems XI; Rome, Italy. 2015.

[123] Hwang J, Jung M, Madapana N, et al. Achieving "synergy" in cognitive behavior of humanoids via deep learning of dynamic visuo-motor-attentional coordination. Humanoid Robots (Humanoids), 2015 IEEE-RAS 15th International Conference on; Seoul. 2015. p. 817-824.

[124] Hochreiter S, Schmidhuber J. Long short-term memory. Neural Computation, 1997;9(8):1735-1780.

[125] Zhang T, Kahn G, Levine S, et al. Learning deep control policies for autonomous aerial vehicles with MPC-guided policy search. 2015. Available: http://arxiv.org/abs/1509.06791

[126] Lillicrap TP, Hunt JJ, Pritzel A, et al. Continuous control with deep reinforcement learning. 2015. Available: http://arxiv.org/abs/1509.02971

[127] Finn C, Levine S, Abbeel P. Guided cost learning: deep inverse optimal control via policy optimization. 2016. Available: http://arxiv.org/abs/1603.00448

[128] Levine S, Finn C, Darrell T, et al. End-to-end training of deep visuomotor policies. J Mach Learning Research. 2016;17:1-40.

[129] Rosenstein M, Barto A.(2004). J. 4 supervised actor-critic reinforcement learning. Handbook of Learning and Approximate dynamic programming, 2, 359.

[130] Houthooft R, Chen X, Duan Y, Schulman J, De Turck F, Abbeel P. (2016). VIME: variational information maximizing exploration. In Advances in Neural Information Processing Systems (pp. 1109-1117).

[131] Osband I, Blundell C, Pritzel A, Van Roy B. (2016). Deep exploration via bootstrapped DQN. In Advances in Neural Information Processing Systems (pp. 4026-4034).

[132] Pinto L, Gupta A. Supersizing self-supervision: learning to grasp from 50k tries and 700 robot hours. 2015. Available: http://arxiv.org/abs/1509.06825

[133] Kappler D, Bohg J, Schaal S. Leveraging big data for grasp planning. 2015 IEEE International Conference on Robotics and Automation (ICRA); Seattle, WA. 2015. p. 4304-4311.

[134] Pratt GA. Is a cambrian explosion coming for robotics? Journal of Economic Perspectives. 2015;29(3):51-60.

[135] Szegedy C, Zaremba W, Sutskever I, et al. Intriguing properties of neural networks. 2013. Available: http://arxiv.org/abs/1312.6199